\documentclass[runningheads]{llncs}
\usepackage[T1]{fontenc}
\usepackage{booktabs}               
\usepackage{subcaption}        
\usepackage{multirow}
\usepackage{caption}   
\usepackage{graphicx,verbatim}
\usepackage{amssymb}
\usepackage{amsmath}
\usepackage{pifont}
\usepackage{hyperref}
\usepackage{color}

\usepackage{bbding}        % 提供 \Envelope

\begin{document}
\title{Hypergraph Tversky-Aware Domain Incremental Learning for Brain Tumor Segmentation with Missing Modalities}
\author{Junze Wang\inst{1\dagger}%
  %\thanks{Equal Contributions}
% index{Wang, Junze}
%
\and
Lei Fan\inst{2,3\dagger}
% index{Fan, Lei}
\and
Weipeng Jing\inst{1} 
% index{Jing, Weipeng}
\and
Donglin Di\inst{4} 
% index{Di, Donglin}
\and
Yang Song\inst{3} 
% index{Song, Yang}
\and
Sidong Liu\inst{5} 
% index{Liu, Sidong}
\and
Cong Cong\inst{5}\Envelope
% index{Cong, Cong}
}

\authorrunning{J. Wang et al.}

\institute{College of Computer and Control Engineering, Northeast Forestry University, Harbin, China 
\and
The Centre for Healthy Brain Ageing (CHeBA), UNSW, Sydney, Australia
\and
School of Computer Science and Engineering, UNSW, Sydney, Australia
\and
School of Software, Tsinghua University, Beijing, China
\and
Centre for Health Informatics, Macquarie University, Sydney, Australia \\
\email{thomas.cong@mq.edu.au}
}

\maketitle              % typeset the header of the contribution
\begingroup                      
  \renewcommand\thefootnote{$\dagger$}
  \footnotetext{Equal contribution}                   
  \renewcommand\thefootnote{\Envelope}
  \footnotetext{Corresponding author}
\endgroup

\begin{abstract}
Existing methods for multimodal MRI segmentation with missing modalities typically assume that all MRI modalities are available during training. However, in clinical practice, some modalities may be missing due to the sequential nature of MRI acquisition, leading to performance degradation. Furthermore, retraining models to accommodate newly available modalities can be inefficient and may cause overfitting, potentially compromising previously learned knowledge.
To address these challenges, we propose Replay-based Hypergraph Domain Incremental Learning (ReHyDIL) for brain tumor segmentation with missing modalities. ReHyDIL leverages Domain Incremental Learning (DIL) to enable the segmentation model to learn from newly acquired MRI modalities without forgetting previously learned information.
To enhance segmentation performance across diverse patient scenarios, we introduce the Cross-Patient Hypergraph Segmentation Network (CHSNet), which utilizes hypergraphs to capture high-order associations between patients. Additionally, we incorporate Tversky-Aware Contrastive (TAC) loss to effectively mitigate information imbalance both across and within different modalities.
Extensive experiments on the BraTS2019 dataset demonstrate that ReHyDIL outperforms state-of-the-art methods, achieving an improvement of over 2\% in the Dice Similarity Coefficient across various tumor regions. Our code is available at \href{https://github.com/reeive/ReHyDIL}{ReHyDIL}.

% Authors must provide keywords and are not allowed to remove this Keyword section.
\end{abstract}
\section{Introduction}
In clinical practice, brain tumor diagnosis is typically supported by observations using four routine Magnetic Resonance Imaging (MRI) modalities: T1, T1CE, T2, and FLAIR \cite{25}. 
Each modality provides unique tissue contrasts that aid in identifying different tumor regions, such as the whole tumor (WT), tumor core (TC), and enhancing tumor (ET) \cite{24}. 
Consequently, accurate and automated segmentation of these modalities is crucial for effective diagnosis and treatment. \par
Recent advances in deep learning have significantly enhanced multimodal brain tumor segmentation \cite{50,51}. However, model performance often degrades due to missing modalities, which is a common issue caused by clinical constraints such as varying acquisition protocols \cite{2}. Existing methods address this issue primarily through feature fusion \cite{3,34,45,46}, knowledge distillation \cite{4,5,47}, and modality masking \cite{6,7,35}. However, these methods often assume that all MRI modalities are available during training, which is a condition often unmet in practice. 
To address this challenge, we draw inspiration from Domain Incremental Learning (DIL) \cite{9,14}. However, directly applying DIL presents challenges.
Specifically, catastrophic forgetting occurs when newly added modalities overshadow previously acquired ones during training, leading to an inter-modality imbalance. Moreover, tumor regions typically occupy only a small portion of the image relative to the background, resulting in an intra-modality imbalance \cite{49,48}. These issues hinder the model’s ability to learn robust cross-modal features, ultimately degrading segmentation performance in missing modality scenarios. \par
Accordingly, we introduce the Replay-based Hypergraph Domain Incremental Learning (ReHyDIL). 
ReHyDIL employs a replay buffer to retain data from previous modalities, enabling the model to continue domain adaptation \cite{53,54}, learn cross-modal features and mitigate catastrophic forgetting.
To address inter-modality imbalance, we propose the Tversky-Aware Contrastive (TAC) loss with a balanced queue. This loss encourages the model to learn robust correlations between different modalities, bridging the information gap caused by missing modalities during inference and reducing modality imbalance.
For intra-modality imbalance at the pixel level, we employ Tversky loss and Tversky-based Focal loss, ensuring better segmentation of small and less-represented tumor structures.
Moreover, to effectively leverage the rich patient-level information in MRI datasets, we propose CHSNet, a hypergraph-based segmentation network designed to capture complex inter-patient relationships. In CHSNet, each patient is represented as a vertex, connected by hyperedges to others with similar tumor features. The contributions can be summarized as follows: \\
\begin{itemize}
    \item We propose ReHyDIL for brain tumor segmentation with missing modality in scenarios where MRI modalities are incrementally available during training. 
    \item We introduce the Tversky-Aware Contrastive (TAC) loss to balance the learning between different modalities.
    \item We present the Cross-Patient Hypergraph Segmentation Network (CHSNet), which captures high-order associations across patients, enhancing feature representations and improving segmentation performance.
    \item Compared to existing state-of-the-art methods, ReHyDIL achieves higher Dice Similarity Coefficients (DSC) on the BraTS2019 dataset, with gains of 1.05\% for Whole Tumor (WT), 2.53\% for Tumor Core (TC), and 1.52\% for Enhancing Tumor (ET).
\end{itemize}

\begin{figure}[!t]
    \centering
    \includegraphics[width=1\linewidth]{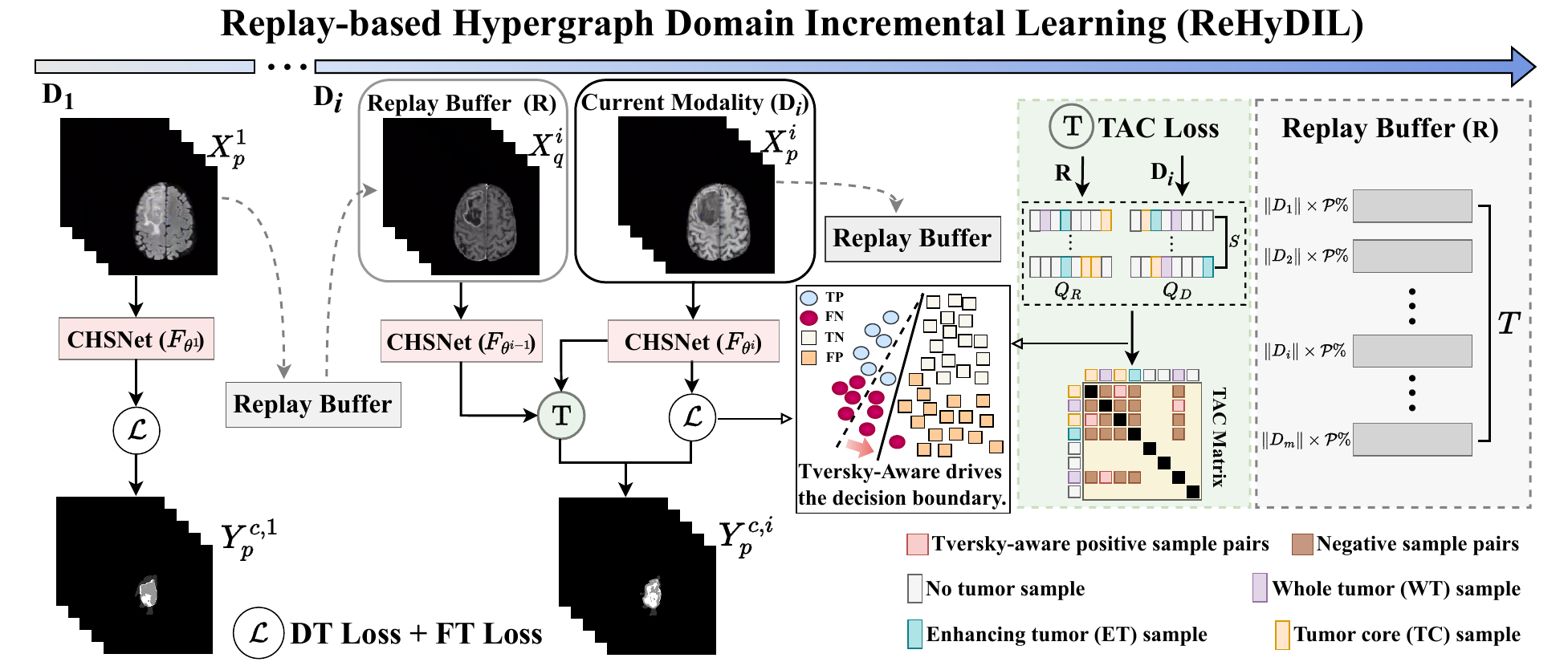}
    \caption{The ReHyDIL pipeline processes a set of MRI modalities acquired sequentially, leveraging incremental learning to train CHSNet. This training is guided by our proposed TAC loss, alongside other Tversky-based segmentation losses. At the end of each learning phase, a replay buffer is constructed to store samples from the currently learned modalities. This mechanism enables the trained CHSNet to effectively capture cross-modality features, improving the segmentation of different tumor regions even in the presence of missing modalities.}
    \label{framework}
\end{figure}

\section{Method} 
\subsection{Replay-based Hypergraph Domain Incremental Learning}
We illustrate the overall workflow of ReHyDIL in Fig. \ref{framework}, where the core objective is to train the proposed CHSNet (Section \ref{sec:net}) $F_{\theta^m}: X_p^m \rightarrow Y_p^{c,m}$ with $m$ continuously arriving MRI modalities. In this study, each consecutively arriving MRI modality strictly corresponds to one training stage. $X_p^m$ and $Y_p^{c,m}$ denote the input sample and its prediction for patient $p$, and $c$ denotes the tumor region label. Initially, $F_{\theta^1}$ is trained using samples from the first modality $D_1$, and a replay buffer is initialized. 
In the subsequent training stages ($1 < i \leq m$), where new modalities are introduced, $F_{\theta^i}$ loads the weights from the previous stage and is trained using the proposed TAC loss (Section \ref{sec:loss}) along with samples from previously encountered modalities stored in the replay buffer.

To build the replay buffer \( R \) with a capacity \( T \), we retain \( \mathcal{P}\% \) of samples from each training stage \( D_i \). Then \( T \) is calculated as $T = \sum_{i=1}^m \| D_{i} \| \times \mathcal{P}\%$. 
For a given modality \( D_i \), we follow the mean loss strategy \cite{41} to select the top \( \mathcal{P}\% \) of samples with loss values closest to the median loss \( \mu \). Specifically, for each sample $X_p^i$, we compute its deviation: $d_p = \left| \mathcal{L}_{total}(X_p^i, \theta^i) - \mu \right|$. Samples are ranked in ascending order based on \( d_p \), with \( \sigma(X_p^i) \) denoting the rank of \( X_p^i \). 
The top \( \mathcal{P}\% \) ranked samples are stored in the replay buffer according to:
\begin{equation}
R = \{X_p^i \in D_i \mid \sigma(X_p^i) \leq  \| D_{i} \| \times \mathcal{P}\% \}.
\end{equation}
Samples stored in $R$ will be used together with $D_i$ to train CHSNet using TAC loss. This enables the final model, $F_{\theta^m}$, to incrementally learn from all modalities and effectively perform segmentation even in the presence of missing modalities.

\subsection{Cross-Patient Hypergraph Segmentation Network}
\label{sec:net}
\begin{figure}[!t]
    \centering
    \includegraphics[width=1\linewidth]{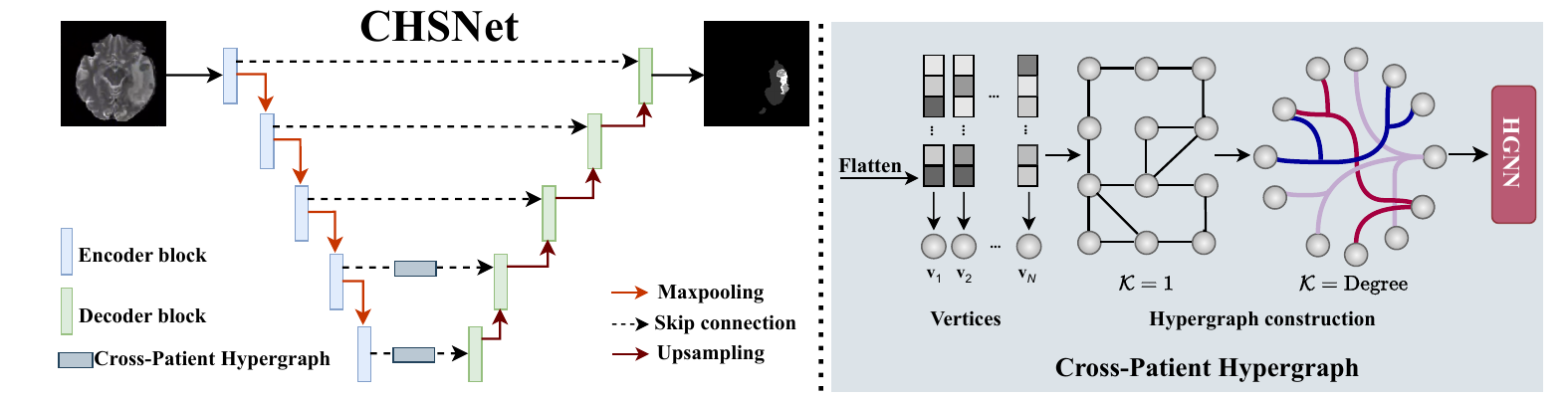}
    \caption{Illustration of the proposed CHSNet. CHSNet follows a U-Net architecture, incorporating hypergraphs into the last and second-to-last encoder layers. These hypergraphs effectively encode cross-patient information into the output features, which serve as inputs to the subsequent decoder blocks.}
    \label{c3h}
\end{figure}
CHSNet is designed to capture cross-patient associations and enhance segmentation performance by integrating a Cross-Patient Hypergraph (CPH). To enforce this cross-patient learning, we impose a constraint during each batch sampling to ensure that selected samples originate from different patients.
As illustrated in Fig. \ref{c3h}, CHSNet is built upon the U-Net architecture \cite{32}. Empirically, we embed the hypergraph into the 4th and 5th encoders to learn different feature representations, enabling more robust segmentation performance.

We construct each hypergraph using the encoder’s output, \( \mathbf{E}_f \in \mathbb{R}^{B \times C \times H \times W} \). First, we flatten \( \mathbf{E}_f \) from \( (B, C, H, W) \) to \( (N, C) \), where \( N = B \times H \times W \) represents the total number of hypergraph vertices. Each vertex corresponds to a \( C \)-dimensional feature vector, formally denoted as: 
$\mathbf{V} = [\mathbf{v}_1;\mathbf{v}_2;\dots;\mathbf{v}_N] \in \mathbb{R}^{N \times C}$, where $\mathbf{v}_j \in \mathbb{R}^{C}$.
To construct the hypergraph \( \mathbf{G} \), we employ a \( \mathbf{K} \)-\( \mathbf{N} \)earest \( \mathbf{N} \)eighbours (\( \mathbf{KNN} \)) method based on Euclidean distance \cite{40,52}. Initially, a \( \mathbf{KNN} \) search is performed for each vertex with \( \mathbf{K} = 1 \), generating a preliminary incidence matrix \( \mathbf{H} \), where a value of 1 indicates a connection between vertices via a hyperedge.  
Next, we refine the hypergraph structure by updating the \( \mathbf{K} \)-value of each vertex \( \mathbf{v}_j \) to match its degree in \( \mathbf{H} \), i.e., \( \mathbf{K}(j) \) is set to the degree of \( \mathbf{v}_j \). A second \( \mathbf{KNN} \) search is then performed using the updated \( \mathbf{K} \)-values to establish the final hypergraph structure. As a result, each vertex \( \mathbf{v}_j \) is connected to its set of nearest neighbors, \( \mathbf{N}_{\mathbf{K}(j)} \), defining the hyperedge as: $\mathbf{e}_j = \{\mathbf{v}_j\} \cup \mathbf{N}_{\mathbf{K}(j)} (\mathbf{v}_j)$.
This process yields the updated incidence matrix \( \mathbf{H} \in \mathbb{R}^{N \times \mathbf{K}} \), along with the vertex degree matrix \( \mathbf{D}_v \) and hyperedge degree matrix \( \mathbf{D}_e \).  

Subsequently, HGNN \cite{4} is employed to integrate cross-patient higher-order relationships. The updated hypergraph features \( \mathbf{V}' \) are computed as follows:
\begin{equation}
\mathbf{V}' = \mathbf{D}_v^{-\frac{1}{2}} \mathbf{H} \mathbf{W}_e \mathbf{D}_e^{-1} \mathbf{H}^\top \mathbf{D}_v^{-\frac{1}{2}} \mathbf{V}, 
\end{equation}
where \( \mathbf{D}_v^{-\frac{1}{2}} \) normalizes vertex degrees, \( \mathbf{W}_e \) is a learnable weight for hyperedges and is initially set to 1.  
Next, we reshape \( \mathbf{V}' \) back to \( (B, C, H, W) \), aligning it with the original feature map \( \mathbf{E}_f \). Finally, we concatenate \( \mathbf{V}' \) and \( \mathbf{E}_f \) along the channel dimension and fuse the resulting features via a \( 1\times1 \) convolution, producing integrated feature representations to be passed into subsequent decoder blocks.

\subsection{Tversky-Aware Contrastive Loss}
\label{sec:loss}
We propose the TAC loss to regularize CHSNet by leveraging samples stored in \( R \) alongside those from the current modality \( D_i \). However, due to the limited size of \( R \), which can be easily overwhelmed by \( D_i \), we introduce a balanced queue (\( Q \)) mechanism to effectively address this inter-modality imbalance. 
\( Q \) is designed as a double-queue structure, where one queue (\( Q_R \)) stores predictions of samples from \( R \) generated by $F_{\theta^{i-1}}$, and the other (\( Q_D \)) contains predictions of samples from \( D_i \) produced by $F_{\theta^i}$. 

Specifically, given a batch of \( B \) predictions, we randomly select \( S \) predictions, where \( S = \min(B, \| R \|) \), from each $Q_R$ and $Q_D$, and normalize them before storing them in the memory bank \( Q \), ensuring that \( \| Q \| = 2S \). 
For each prediction \( Y_p^c \in Q_{i} \), we define its positive pair \( Y_q^c \in Q_{j\neq i} \). This means that they come from different patients and different modalities but share the same predicted region. Similarly, we define the negative pair \( Y_q^z \in Q_{j\neq i} \), where \( p \neq q \) and \( c \neq z \). Then contrastive learning is conducted via:
\begin{equation}
\mathcal{L}_{\mathrm{TAC}}=-\log \frac{\exp(S_{\mathrm{tve}}
(Y_p^c, Y_q^c) / \tau)}{
  \exp(S_{\mathrm{tve}}(Y_p^c, Y_q^c) / \tau)+
  \sum_{p \neq q, c \neq z} \exp(S_{\mathrm{tve}}(Y_p^c, Y_q^z) / \tau)},
\label{eq3:tac}
\end{equation}
where $\tau$ is the temperature coefficient and is set to 1, $S_{\mathrm{tve}}$ is the Tversky similarity. Following \cite{42}, we define $S_{\mathrm{tve}}$ as:
\begin{equation}
S_{\mathrm{tve}}(g, u)=\frac{gu}{gu + \alpha g(1-u) + \beta(1 - g) u},
\label{eqs:ts}
\end{equation}
where \( g \) and \( u \) denote two different predictions, \( \alpha \) and \( \beta \) are hyperparameters that control the trade-off between false positives (FPs, i.e., false detections) and false negatives (FNs, i.e., missed detections). In Equation~\ref{eq3:tac}, the anchor prediction \( Y_{p}^{c} \) is used as \( g \), while the predictions of the paired samples \( Y_{q}^{c} \), \( Y_{q}^{z} \) are used as \( u \). 

However, due to intra-modality imbalance, a naively trained CHSNet may produce high-precision but low-recall segmentations, where \( \text{FNs} \gg \text{FPs} \). To mitigate this imbalance, we explicitly set $\beta > \alpha$, applying a greater penalty to FNs to better capture rare positive objects, as shown in Fig. \ref{framework}. 
Furthermore, we design an intra-modality regularization $\mathcal{L}_\mathrm{intra} = \mathcal{L}_\mathrm{DT} + \mathcal{L}_\mathrm{FT}$ under this setup ($\beta > \alpha$).
Specifically, $\mathcal{L}_\mathrm{DT}$ is a Tversky-based Dice loss, which extends Dice loss with Tversky similarity to systematically suppress FNs. Furthermore, $\mathcal{L}_\mathrm{FT}$ integrates Tversky similarity into Focal loss \cite{43} to further reduce FNs by focusing on difficult-to-classify samples:
\begin{equation}
\mathcal{L}_\mathrm{DT}(y,t) = 1- \sum_{i=1}^n S_\mathrm{tve}(y_i,t_i), \quad
\mathcal{L}_\mathrm{FT}(y,t) = (1- \sum_{i=1}^n S_\mathrm{tve}(y_i,t_i))^{\gamma},
\label{eq5}
\end{equation}
where $y_i$ denotes the $i$-th predicted pixel probability, $t$ denotes its ground truth label, and $\gamma$ is used to adjust the level of attention given to minority class pixels.
The total loss is supervised by $\mathcal{L}_\mathrm{TAC}$, and $\mathcal{L}_\mathrm{intra}$:
\begin{equation}
\mathcal{L}_\mathrm{total} = \omega \mathcal{L}_\mathrm{TAC} + \mathcal{L}_\mathrm{intra}.
\label{eq7}
\end{equation}

\begin{table}[!t]
% \small T1c and FL indicate T1CE and Flair, respectively. The highest scores are highlighted in bold.
% \scriptsize
\fontsize{10}{10}\selectfont
\raggedright
\caption{Segmentation results of WT, TC, and ET in terms of DSC(\%) on the BraTS2019 dataset. Missing and available modalities are denoted by $\circ$ and $\bullet$, respectively. “$\dagger$ ”denotes significance level is reached as p-value $<$ 0.05.}
\label{tab1}
\scalebox{0.7}{
\renewcommand{\arraystretch}{1.3} 
\begin{tabular}{cccc|ccccc|ccccc|ccccc}
\cline{1-19}
\multicolumn{4}{c|}{Modalities} & \multicolumn{5}{c|}{WT} & \multicolumn{5}{c|}{TC} & \multicolumn{5}{c}{ET}\\
\cline{1-19}
T2 & T1c & T1 & FL & mmF & MFI  & M$^3$ & M$+$P & Ours & mmF & MFI  & M$^3$ & M$+$P & Ours &  mmF & MFI  & M$^3$ & M$+$P & Ours \\
\cline{1-19}
$\bullet$ & $\circ$ & $\circ$ & $\circ$  & 80.66 & 81.87 & 84.49 & 82.74 & \textbf{86.71}$^\dagger$ & 69.45 & 58.26 & 71.35 & 55.74 & \textbf{73.68}$^\dagger$ & 37.93 & 39.82 & 47.61 & 36.29 & \textbf{49.75}$^\dagger$\\
$\circ$ & $\bullet$ & $\circ$ & $\circ$  & 73.64 & 74.39 & 78.62 & 75.88 & \textbf{79.48}$^\dagger$ & 78.62 & 81.32 & 83.95 & 82.83 & \textbf{84.65}$^\dagger$ & 70.64 & 76.53 & 79.33 & 79.29 & \textbf{80.17}$^\dagger$ \\
$\circ$ & $\circ$ & $\bullet$ & $\circ$  & 73.76 & 73.28 & 79.32 & 78.32 & \textbf{80.77}$^\dagger$ & 61.36 & 65.14 & 65.98 & 62.13 & \textbf{69.34}$^\dagger$ & 33.05 & 33.82 & 44.43 & 38.40 & \textbf{45.22}$^\dagger$ \\
$\circ$ & $\circ$ & $\circ$ & $\bullet$  & 87.31 & 86.96 & 86.38 & 87.05 & \textbf{88.25}$^\dagger$ & 66.85 & 57.31 & 70.37 & 64.53 & \textbf{72.87}$^\dagger$ & 37.37 & 40.67 & 45.62 & 38.23 & \textbf{48.61}$^\dagger$ \\
$\bullet$ & $\bullet$ & $\circ$ & $\circ$  & 84.55 & 85.81 & 85.74 & 84.47 & \textbf{87.13}$^\dagger$ & 79.98 & 85.20 & 83.25 & 82.94 & \textbf{85.96}$^\dagger$ & 71.92 & 80.04 & 78.76 & 79.75 & \textbf{80.51}$^\dagger$ \\
$\circ$ & $\bullet$ & $\bullet$ & $\circ$  & 77.60 & 74.21 & 80.31 & 79.95 & \textbf{81.48}$^\dagger$ & 79.98 & 83.09 & 85.43 & \textbf{86.59} & 86.32  & 71.38 & 77.47 & 80.31 & \textbf{82.35} & 80.97 \\
$\circ$ & $\circ$ & $\bullet$ & $\bullet$  & 88.90 & 88.09 & 90.17 & 89.12 & \textbf{89.63}$^\dagger$ & 70.44 & 65.96 & 66.32 & 66.59 & \textbf{71.75}$^\dagger$ & 41.93 & 41.47 & 48.83 & 42.49 & \textbf{49.55}$^\dagger$ \\
$\bullet$ & $\circ$ & $\bullet$ & $\circ$  & 83.17 & 84.52 & 86.29 & 85.30 & \textbf{87.58}$^\dagger$ & 68.73 & 60.63 & 65.52 & 62.16 & \textbf{69.42}$^\dagger$ & 42.46 & 43.36 & 44.65 & 38.98 & \textbf{47.36}$^\dagger$ \\
$\bullet$ & $\circ$ & $\circ$ & $\bullet$  & 88.68 & 89.53 & 90.21 & 88.67 & \textbf{90.39} & 70.72 & 65.43 & 70.21 & 68.48 & \textbf{70.96}$^\dagger$ &  43.54 & \textbf{47.29} & 46.69 & 46.75 & 46.93 \\
$\circ$ & $\bullet$ & $\circ$ & $\bullet$  & 88.50 & 89.91 & 90.75 & 89.56 & \textbf{91.42}$^\dagger$ & 80.84 & 82.26 & 83.02 & 82.37 & \textbf{83.22}$^\dagger$ &  72.03 & 77.52 & 78.17 & 75.58 & \textbf{78.65}$^\dagger$ \\
$\circ$ & $\bullet$ & $\bullet$ & $\bullet$  & 89.02 & 89.78 & 89.89 & 89.91 & \textbf{90.75}$^\dagger$ & 82.33 & 85.96 & 86.13 & 86.06 & \textbf{86.46} & 72.82  & 83.23 & \textbf{84.17} & 83.19 & 83.26  \\
$\bullet$ & $\circ$ & $\bullet$ & $\bullet$  & 89.08 & 90.06 & 90.44 & 89.99 & \textbf{91.19}$^\dagger$ & 71.90 & 66.37 & 71.87 & 59.87 & \textbf{72.50}$^\dagger$ & 45.34 & 47.89 & 50.26 & 37.08 & \textbf{48.11}$^\dagger$  \\
$\bullet$ & $\bullet$ & $\circ$ & $\bullet$  & 88.94 & 91.05 & 91.21 & 89.83 & \textbf{91.46}$^\dagger$ & 81.64 & 86.99 & 86.53 & 86.20 & \textbf{87.79}$^\dagger$ & 72.25 & 83.32 & 82.48 & 82.10 & \textbf{84.62}$^\dagger$ \\
$\bullet$ & $\bullet$ & $\bullet$ & $\circ$  & 84.95 & 86.47 & 87.89 & 86.59 & \textbf{87.98}$^\dagger$ & 81.27 & 85.58 & 86.41 & 88.12 & \textbf{89.24}$^\dagger$ & 72.98 & 81.11 & 82.35 & 82.88 & \textbf{83.46}$^\dagger$ \\
$\bullet$ & $\bullet$ & $\bullet$ & $\bullet$  & 89.21 & 90.81 & 91.36 & 90.35 & \textbf{92.43}$^\dagger$ & 82.47 & 86.79 & 86.51 & 87.73 & \textbf{88.06}$^\dagger$ & 72.74 & 83.34 & 83.88 & 82.76 & \textbf{85.35}$^\dagger$ \\
\cline{1-19}
\multicolumn{4}{c|}{Means} & 84.53 & 85.12 & 86.87 & 85.91 & \textbf{87.78} & 75.09 & 74.42 & 77.53 & 74.82 & \textbf{79.49} & 57.23 & 62.46 & 65.17 & 61.74 & \textbf{66.16} \\
\cline{1-19}

\end{tabular}
}
\end{table}
% As illustrated in Table \ref{tab1}, the performance of these methods is listed for three distinct tumor types on all 15 possible cases of missing modality. 

\section{Experiment}
\textbf{Datasets.}
BraTS2019 dataset \cite{26} is a multimodal MRI dataset specifically designed for brain tumor segmentation, comprising 335 patient-level 3D image volumes. Each case includes four modalities: T1, T1CE (T1c), T2, and FLAIR (FL), and three tumor regions: ET, TC, and WT. For each 3D volume, 2D slices were extracted along the z-axis and cropped to a resolution of 224 × 224 pixels. The dataset was divided into training, validation, and testing subsets using an 8:1:1 ratio, resulting in 41,540 training slices from 268 patients, 5,115 validation slices from 33 patients, and 5,270 testing slices from 34 patients. To maintain data integrity, slices from the same patient were assigned to the same subset. 

\noindent \textbf{Experimental Setup.}
All models were trained using PyTorch on 2 $\times$ NVIDIA Tesla V100 GPUs for 200 epochs with a batch size of 64. The Adam optimizer was employed with a learning rate of $1 \times 10^{-4}$ and a weight decay of $4 \times 10^{-4}$. A warm-up multi-step learning rate scheduler with four steps was utilised. For hyperparameters, we set $\omega = 0$ for $D_1$; for all subsequent modalities, $\omega$ was set to 1. We set $\mathcal{P} = 10$, $\alpha = 0.7$, $\beta = 1.5$, and $\gamma=1.2$. 
Furthermore, the modalities were utilized sequentially for training, following the typical order of clinical acquisition \cite{39}: T1$\rightarrow$T2$\rightarrow$FLAIR$\rightarrow$T1CE. Missing modality scenarios were implemented by setting $D_i$ (where $i \in \{\mathrm{T2, T1CE, T1, FLAIR}\}$) to 0. For the evaluation metric, we calculated the Dice Similarity Coefficient (DSC) for each tumor region within each patient's 3D image volume and compared our method with the second-best method using a paired t-test to obtain p-values. 
% The reported results are the average DSCs across all patients.

\begin{figure}[!t]
    \centering
    \includegraphics[width=1\linewidth]{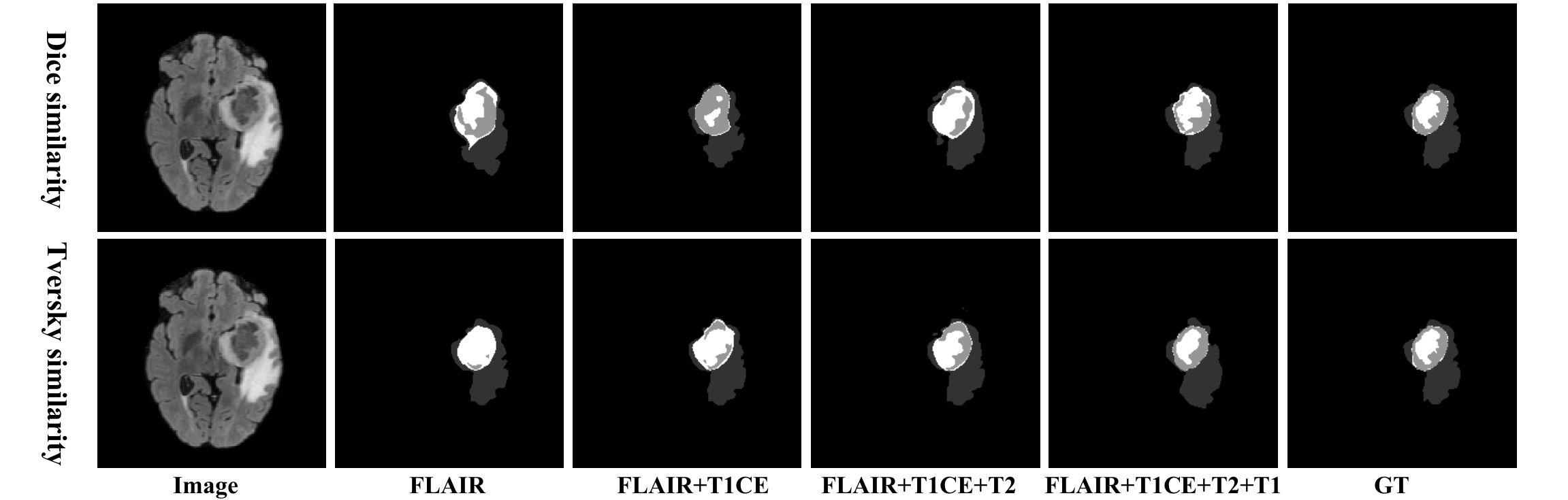}
    \caption{The segmentation visualizations of Dice-based method and ReHyDIL (Tversky similarity), where the difference is only the similarity used. WT, TC, and ET are marked with light gray, medium gray and white, respectively.}
    \label{vis}
\end{figure}

\noindent \textbf{Quantitative Results.} 
We compared the proposed method with state-of-the-art approaches for missing modality brain tumor segmentation, including feature-fusion methods mmFormer (mmF) \cite{3} and UNet-MFI (MFI) \cite{34}, knowledge distillation methods M$^3$AE (M$^3$) \cite{4} and PASSION \cite{5}, and the modality-masked method M$^2$FTrans (M$+$P) \cite{35}. Specifically, PASSION is applied to M$^2$FTrans (M$+$P) as described in \cite{5}. 
As shown in Table \ref{tab1}, ReHyDIL outperforms all compared methods in 41 out of 45 cases, demonstrating a superior average Dice Similarity Coefficient (DSC). Specifically, for Whole Tumor (WT), ReHyDIL achieves an average DSC of 87.78\%, representing a 3.13\% improvement over MFI \cite{34}. For Tumor Core (TC), it attains an average DSC of 79.49\%, reflecting a 2.53\% gain relative to M$^3$ \cite{4}. Regarding Enhancing Tumor (ET), ReHyDIL reaches an average DSC of 66.16\%, signifying a 7.16\% enhancement over M$+$P \cite{35}. Moreover, the p-values are less than 0.05 in 39 out of 45 cases, indicating that the performance improvements of ReHyDIL over other baselines are statistically significant.
It is noteworthy that all compared methods assume the availability of all MRI modalities during training. In contrast, our method operates under a more realistic scenario where modalities are obtained incrementally. The superior performance of ReHyDIL further demonstrates its potential as a reliable tool for real clinical applications.

\noindent \textbf{Qualitative Results.} 
The segmentation visualizations are presented in Fig. \ref{vis}. Integrating additional modalities provides complementary features, which enhance segmentation performance. For example, when only the FLAIR modality is available, ReHyDIL can roughly delineate the whole tumor (WT) but struggles to accurately segment the enhancing tumor (ET) and tumor core (TC). Incorporating T1CE and T2 modalities (i.e., FLAIR+T1CE+T2) significantly improves ET segmentation, achieving a closer alignment with the ground truth (GT).
Furthermore, the visualizations demonstrate that Tversky similarity outperforms Dice similarity in segmenting boundary regions and smaller regions, highlighting its effectiveness in mitigating intra-modality imbalance.

\noindent \textbf{Ablation Study.}
We first compared the impact of adding Cross-Patient Hypergraph (CPH) modules to different layers of CHSNet, as shown in Table \ref{abla}(a). The results indicate that two CPH layers yield the best segmentation performance, while adding a third layer causes a significant decline due to over-smoothing. Additionally, we examined the effectiveness of different similarity metrics in the TAC loss. Compared to cosine similarity \cite{44}, using Tversky similarity improves the Dice score, as it directly measures the overlap between target regions, effectively mitigating intra-modality imbalance.

Furthermore, the ablation studies reveal a key trade-off in controlling false negatives (FNs), as shown in Table \ref{abla}(b). With $\alpha = \beta = 0.5$, Tversky similarity $S_\mathrm{tve}$ simplifies to the Dice similarity, which performs poorly due to an insufficient FN penalty. Gradually increasing $\beta$ from 1.3 to 1.5 significantly improves DSC across all regions, highlighting the importance of penalizing FNs to reduce missed detections. The best performance is achieved at $\alpha =0.7, \beta =1.5$. However, further increasing $\beta$ to 1.6 leads to performance collapse, likely from over-suppressing ambiguous boundaries.
Finally, we conducted ablation studies on the overall framework. Without CPH, CHSNet reduces to the U-Net, and without TAC loss, the model relies solely on intra-modality regularization $\mathcal{L}_\mathrm{intra}$. These results confirm that both CPH and TAC loss individually and collectively enhance segmentation performance—CPH by integrating high-order inter-patient relationships and TAC loss by bridging the information gap during inference.

\begin{table}[t]
    \centering
    % \caption{Ablation study for (a) the encoder layers where CPH is added, (b) different similarity measure in contrastive losses, (c) different modules in ReHyDIL.}
    \caption{Ablation study for (a) the encoder layers where CPH is added and different similarity measure in contrastive losses, (b) Tversky similarity hyperparameters $\alpha$ and $\beta$, (c) different modules in ReHyDIL.}  
    \label{abla}
    % \fontsize{10}{10}\selectfont
    \scalebox{0.7}{
    \begin{subtable}[b]{0.5\linewidth} 
        \raggedright
        \begin{tabular}{cc|ccc}
            \cline{1-5}
            & & \multicolumn{3}{c}{DSC(\%)} \\
            \cline{3-5}
             &  & WT & TC & ET \\
            \cline{1-5}
            % None  & 82.56  & 71.85 & 58.76 \\
            \multirow{3}{*}{Layers}& 5 & 85.32 & 76.48 & 61.95 \\
            & 5, 4 & \textbf{87.78} & \textbf{79.49} & \textbf{66.16} \\
            & 5, 4, 3 & 79.95 & 68.59 & 50.82 \\
            \cline{1-5}
           \multirow{2}{*}{Similarity} & Cosine & 84.18 & 75.03 & 62.74 \\
            & Tversky & \textbf{87.78} & \textbf{79.49} & \textbf{66.16} \\
            \cline{1-5}
        \end{tabular}
        \caption{}   
    \end{subtable}
    \hfill
    %---- (c) ----
    \begin{subtable}[b]{0.35\linewidth} 
        \raggedright
        \begin{tabular}{cc|ccc}
            \cline{1-5}
            \multirow{2}{*}{$\alpha$} & \multirow{2}{*}{$\beta$} & \multicolumn{3}{c}{DSC(\%)} \\
            \cline{3-5}
                 & & WT & TC & ET \\
            \cline{1-5}
            0.5 & 0.5  &  80.12  & 74.58  &  60.31 \\
            0.9 & 1.3  &  84.73 & 77.04  &  63.19 \\             
            0.8 & 1.4  &  85.96 & 77.27  & 64.85  \\
            0.7 & 1.5 & \textbf{87.78} & \textbf{79.49} & \textbf{66.16} \\ 
            0.6 & 1.6  &  85.29 & 76.71  &  62.48 \\
            \cline{1-5}
        \end{tabular}
        \caption{}
    \end{subtable}
    \hfill
    %---- (d) ----
    \begin{subtable}[b]{0.45\linewidth}
        \raggedright     
        \begin{tabular}{ccc|ccc}
            \cline{1-6}
            \multirow{2}{*}{DIL} & \multirow{2}{*}{CPH} & \multirow{2}{*}{TAC} & \multicolumn{3}{c}{DSC(\%)} \\
            \cline{4-6}
              &  &   & WT & TC & ET \\
            \cline{1-6}
            \ding{52} &  &  & 60.14 & 44.62 &  33.58 \\
            \ding{52} & \ding{52} &  & 76.42 & 55.04 &  47.29 \\
            \ding{52} &  & \ding{52} & 82.56  & 71.85 & 58.76   \\
            \ding{52} & \ding{52} & \ding{52} & \textbf{87.78} & \textbf{79.49} & \textbf{66.16} \\
            \cline{1-6}
        \end{tabular}
        \caption{}
    \end{subtable}
    \hfill
    }
\end{table}
\section{Conclusion} 
In this paper, we propose the ReHyDIL framework for brain tumor segmentation with missing modalities, leveraging DIL to incrementally learn from new modalities while preserving prior knowledge.
Specifically, we propose TAC loss, which effectively balances the learning process across different modalities and mitigates the information gap caused by missing modalities during inference. 
Additionally, we introduce CHSNet to capture high-order relationships between patients, enhancing segmentation performance.
Extensive experiments on the BraTS2019 dataset demonstrate that ReHyDIL achieves state-of-the-art performance in missing modality scenarios.
Though effective, ReHyDIL has certain limitations. Its use of a replay buffer introduces additional memory overhead, and the framework is currently limited to four MRI modalities with similar image characteristics. In future work, we aim to explore alternative DIL strategies and develop more advanced alignment mechanisms or adaptors to effectively fuse features from diverse modalities.

\begin{credits}
\subsubsection{\discintname}
The authors have no competing interests to declare that are relevant to the content of this article.
\end{credits}

\bibliographystyle{splncs04}
\bibliography{ref}

\end{document}